# 3D Convolutional Neural Networks for Dendrite Segmentation Using Fine-Tuning and Hyperparameter Optimization

Jim James[a], Nathan Pruyne[a,b], Tiberiu Stan[b], Marcus Schwarting[c,*], Jiwon Yeom[d], Seungbum Hong[d], Peter Voorhees[b], Ben Blaiszik[a,c], Ian Foster[a,c]

[a]*Data Science and Learning Division, Argonne National Laboratory, 9700 Cass Avenue, Lemont, IL 60439*

[b]*Department of Materials Science and Engineering, Northwestern University, 2220 Campus Drive, Cook Hall, Evanston, IL 60208, USA*

[c]*Department of Computer Science, University of Chicago, 5801 South Ellis Avenue, Chicago, IL 60637*

[d]*Department of Materials Science and Engineering, Korea Advanced Institute of Science and Technology, Daejeon 34141, Korea*

[*]Corresponding author email: meschw04@uchicago.edu

## Abstract

Dendritic microstructures are ubiquitous in nature and are the primary solidification morphologies in metallic materials. Techniques such as x-ray computed tomography (XCT) have provided new insights into dendritic phase transformation phenomena. However, manual identification of dendritic morphologies in microscopy data can be both labor intensive and potentially ambiguous. The analysis of 3D datasets is particularly challenging due to their large sizes (terabytes) and the presence of artifacts scattered within the imaged volumes. In this study, we trained 3D convolutional neural networks (CNNs) to segment 3D datasets. Three CNN architectures were investigated, including a new 3D version of FCDense. We show that using hyperparameter optimization (HPO) and fine-tuning techniques, both 2D and 3D CNN architectures can be trained to outperform the previous state of the art. The 3D U-Net architecture trained in this study produced the best segmentations according to quantitative metrics (pixel-wise accuracy of 99.84% and a boundary displacement error of 0.58 pixels), while 3D FCDense produced the smoothest boundaries and best segmentations according to visual inspection. The trained 3D CNNs are able to segment entire 852 x 852 x 250 voxel 3D volumes in only ~60 seconds, thus hastening the progress towards a deeper understanding of phase transformation phenomena such as dendritic solidification.

**Graphical Abstract**

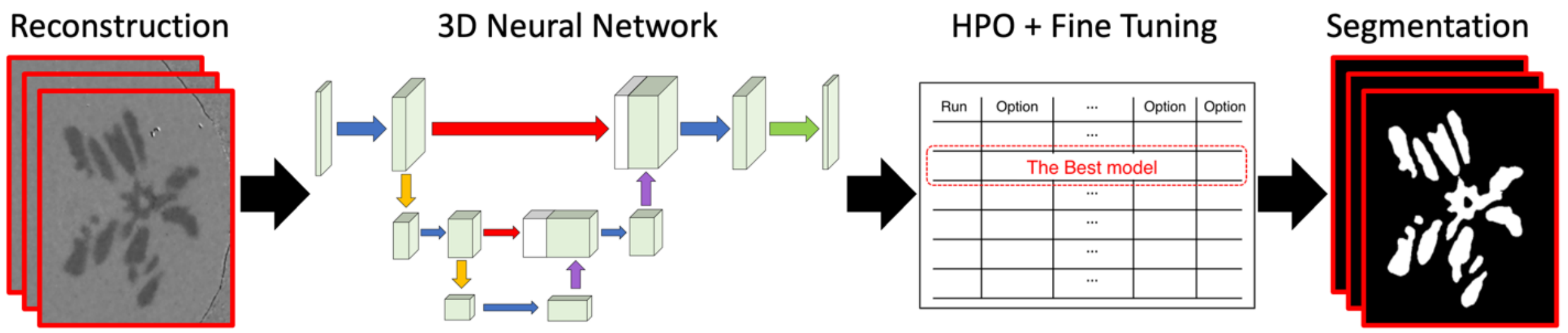

## 1. Introduction

Almost all metallic materials used in society are fabricated using processes that involve solidification [1]. Examples include casting, additive manufacturing, welding, soldering, and brazing. In most cases the metallic alloys solidify through the formation of dendritic microstructures [2]. The morphologies of dendritic crystals impact many mechanical, thermal, electrical, and chemical properties [1,3,4]. Thus, characterizing the formation and evolution of dendritic microstructures is critical for understanding and applying processing-structure-property relationships in materials [5].

Microstructural imaging techniques such as optical microscopy, atomic force microscopy, scanning electron microscopy, serial sectioning, and x-ray computed tomography (XCT) allow researchers to observe and quantitatively characterize such growth processes with unprecedented accuracy. To extract quantitative materials information from these characterization techniques, the images must first be segmented into multiple useful parts. Segmentation is the act of assigning a class to each pixel of an image in order to identify and group features of interest. Manually segmenting image slices of dendritic solidification is impractically time-consuming. A human researcher may use tools such as Photoshop or GIMP to trace dendrite boundaries across a single image in roughly 20 minutes [6]. Thus, a researcher working 40 hours per week might expect to spend eight years manually segmenting a dataset that was collected in a matter of minutes. To overcome this segmentation challenge, materials scientists have turned to machine learning (ML) methods to alleviate the burden [6].

Deep learning methods, especially 2D convolutional neural networks (CNNs), have proven effective for segmentation in both the dendritic context [7] and other materials science applications [8]. These methods are covered in more detail in Section 2. However, analyzing 3D datasets as stacks of 2D images is not ideal because valuable information across slices (i.e. in the z-direction) may be lost while investigating one slice at a time. Problems may also arise when there are z-slice variations within the datasets. For example, XCT datasets often include ring artifacts [7] which alter pixel intensity values and may appear in only some z-slices of the 3D datasets. Furthermore, serial sectioning techniques often leave streaks or speckles on imaged surfaces which vary from slice to slice [7]. These artifacts may hinder the ability to identify features of interest correctly, and thus there is a need for 3D image analysis techniques that take advantage of the full multi-dimensional nature of 3D datasets.

In this paper we showcase how using 3D CNNs along with hyperparameter optimization and fine-tuning produces segmented datasets that are more accurate than the previous 2D state of the art techniques.

## 2. Background

### 2.1. Classical Computer Vision Segmentation Approaches

Classical computer vision and machine learning procedures such as Otsu's method [9], Expectation Maximization/Maximization of Posterior Marginals (EMMPM) [10], k-means clustering [11], and support vector machines (SVMs) [12] are all reasonably performant for simple segmentation tasks. One such successful application of these techniques is optical character recognition [13], as written characters generally exhibit uniform and distinct luminance as well as easily identifiable edges. XCT and serial sectioning datasets rarely exhibit such uniformity, and edges are often difficult to identify. Untrained human observers may incur subjective biases on their respective interpretations of an edge location, making decisive ground truth segmentations elusive to all but domain experts. Figure 1 demonstrates the performance and shortfalls of these methods for a single segmented image overlay for each classical technique. The reconstruction image is taken from a 4D XCT experiment of dendritic solidification. The dark gray patches represent a cross-section through a dendritic solid crystal, while the light gray surrounding areas correspond to the background. The ground truth was obtained using a combination of computational and manual segmentation techniques (details are in Section 3), and is considered as accurate as possible. Figure 1 also presents segmentation overlays from five computational methods. Pixels in black are correctly segmented as background, and pixels in white are correctly segmented as dendrite. Pixels in green represent regions incorrectly classified as dendrite compared to the ground truth (false positives or over-segmentation). Pixels in pink represent regions that was misclassified as background (false negatives or under-segmentation). This convention for overlay labeling is maintained throughout the rest of this paper.

As indicated by the large number of green patches in Figure 1, segmentations using EMMPM, Otsu's Method, and K-means are not of high quality. SVM and 2D SegNet produce more accurate segmentations, but some dendrites are connected where they should not be. Red arrows indicate an area that is particularly challenging for classical computer vision methods to correctly assess due to the diffuse dendrite-background boundaries. It is difficult to identify and correct for areas which should not have been connected, thus it is important that segmentations do not contain these connection errors.

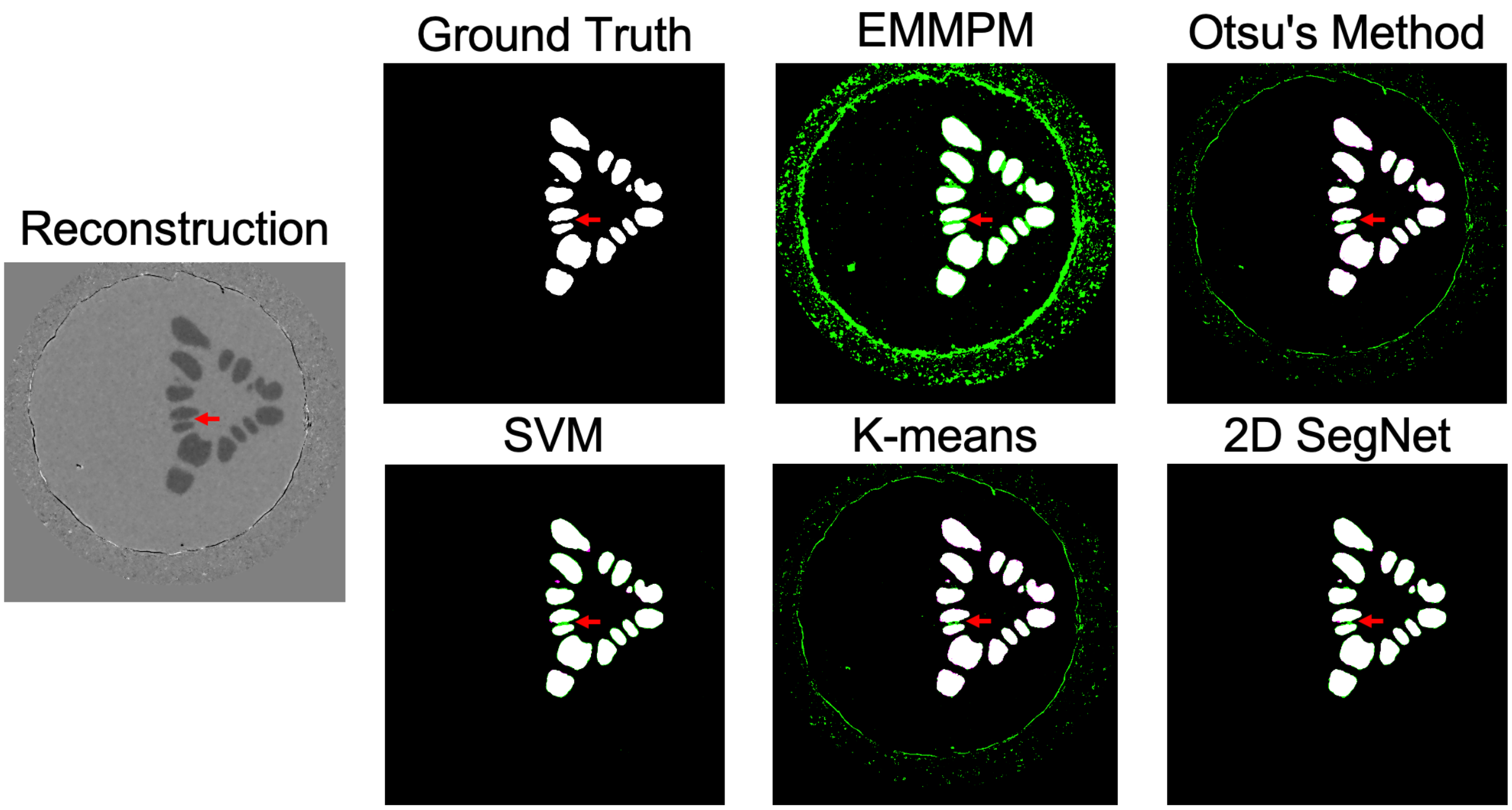

**Figure 1.** Overlays for segmentations using EMMPM, Otsu's method, SVM, K-means, and the 2D SegNet. In these segmentations (and in all others), white represents true positives, black represents true negatives, green represents false positives, and pink represents false negative.

2.2. Deep Learning Architectures and Approaches

CNN architectures were first effectively demonstrated by LeCun et al. for document segmentation, classification, and analysis [14]. In the past two decades, a multitude of CNN-based approaches have helped form the foundation of visual deep learning [15]. In materials science, 2D CNN architectures have been applied for classification [16], segmentation [17], bounding [18], and temporal assessment [19]. Stan et al. demonstrated that a 2D CNN can significantly out-perform classical computer vision approaches on XCT and serial sectioning datasets, enabling a more coherent understanding of dendrite formation in alloys [6].

Typical CNNs contain an encoder consisting of convolutions and pooling layers which encode varying scales of features in the input [20]. To convert these low-resolution encoded features to a segmentation map, neural networks used for segmentation have a decoder which restores the original spatial resolution by performing up-sampling, concatenation with features from the encoder, and additional convolutions to produce a high-resolution segmentation map.

Additionally, Ronneberger et al. [21] developed U-Net, a high performing fully-convolutional network (FCN) with a similar number of filters on the decoding path as the encoding path. A substantial increase in the number of decoding path filters has been shown to preserve contextual information throughout the decoding path. Finally, Jégou et al. [22] designed Fully Convolutional DenseNet (FCDense), an FCN extended from DenseNet. FCDense is built with many short connections between all feature maps, resulting in a more efficient use of training parameters, as well as the improved utilization of features across multiple scales.

In conjunction with substantial performance gains from altering underlying network architectures, additional training techniques have evolved for improving model performance. These include data augmentations, network fine-tuning, and hyperparameter optimization. *Data augmentation* processes involve transforming original training images in multiple ways (eg. blurring, rotating, reflecting, skewing) and setting transformed images alongside original images so that all may be used during training [23]. It has been demonstrated that by selecting and training on a large set of image transforms for data augmentation, a trained model can be made less sensitive to alterations in initial conditions by increasing the variance in training data [24]. *Fine-tuning*, or the process of transferring pre-trained weights to a model trained on a different dataset, is an efficient way to generate networks that are both accurate and versatile to a large amount of data [25]. As observed by Lei et al. [26], fine-tuning of CNNs often improves overall accuracy when labeled data is limited. Finally, *hyperparameter optimization* (HPO) is the process of optimizing network hyperparameters, which are parameters that are not learned by the network architecture itself. It has been demonstrated that model performance is highly dependent on hyperparameters including learning rate, patch width, patch overlap, selection of augmentations, and batch size [27]. These techniques were used in our work to obtain CNNs with superior performance, as reported in the Results section.

## 3. Materials and Methods

### 3.1. Experimental Processes

The training and test data were collected from an in-situ solidification XCT experiment performed at Argonne National Laboratory Advanced Photon Source, beamline 2-BM. A cylindrical section measuring ~1 mm in diameter and ~6 mm in height was first machined from an Al-20wt%Zn alloy. The sample was then fitted into a boron nitride holder and placed into a custom furnace, where the temperature was raised to ~625 °C. After complete melting, the assembly was rotated with a slew speed of two revolutions per second and slowly cooled at a rate of ~1 °C/minute, with tomographic images captured at roughly 2 ms intervals. Any difference in phase composition results in a strong tomographic contrast between the nucleated dendritic solid and the surrounding molten metal. A total of ~115,000 projections were collected over ~3.3 min. The final 3D grayscale volumetric datasets were reconstructed from the projections using the time-interlaced model-based iterative reconstruction (TIMBIR) algorithm [28]. The result is a set of 200 reconstructions, each representing the dendritic microstructure during a 1 s interval. Each 3D reconstruction is a grayscale volumetric dataset of size 852 x 852 x 250 voxels, which can also be represented as a stack of 250 z-slice images of size 852 x 852 pixels. For more information on the experimental process, see Stan et al. [7].

### 3.2. Generating Ground Truth Labels and Datasets

From the reconstructions generated, we created two datasets. The first dataset, "SegNet output”, consists of a total of 1750 z-slice images from seven different reconstructions. The ground truths (labels) for this dataset were generated using the best 2D SegNet CNN model obtained by Stan et al. [7].

The second dataset, “HandSeg”, was generated using a combination of computational and manual segmentation techniques. This dataset consists of 112 images from two reconstructions. Each image was labelled using methods described in reference [6]. The images were first computationally blurred to soften any sharp features, then a Canny Edge Detection algorithm was applied to identify most of the dendrite edges. Each image was then manually edited using the GNU Image Manipulation Program (GIMP), and verified by the authors. Finally, the segmented image stacks were computationally smoothed using 3D erosion and dilation operations in MATLAB to account for inconsistencies in the z-direction, and verified again by the authors. This extensive procedure produced segmentations that we consider to be as accurate as reasonably possible.

For the purposes of neural network model training, all datasets were split into train, test, and validation sets. For the SegNet output dataset, volumes of one or four z-slices (depending on the model) were randomly sorted such that 70% of the volumes were used for training, 20% for validation, and 10% for testing. For the HandSeg dataset, we created 14 groups each consisting of eight labeled images. We then designated six of these groups for training, four for validation, and four for testing. All segmentation metrics reported in this paper were measured using the four testing groups from the HandSeg dataset (totaling 32 images).

By using two different datasets, we ensured that the models trained using both datasets were applicable to a wide variety of data while maintaining high accuracy. The wide breadth of data in the SegNet dataset ensured that the model will not be overtrained to a small number of images, while the high accuracy of the HandSeg dataset helped increase model accuracy. We leveraged the strength of both datasets via fine-tuning, which we describe in greater detail in section 3.3.

### 3.3. Deep Learning Implementations

We worked with one 2D CNN and two 3D CNN architectures in this study. The 2D CNN and the first 3D CNN are U-Net architectures implemented in [29], with the 3D implementation developed by Cicek et al. [30]. The second 3D CNN is 3D FCDense [22]. Both 3D CNNs have a similar network design, as shown in Figure 2a. The colored arrows represent operations that are performed as the 3D volumetric inputs are passed through the network. The final output is a volume where each voxel contains a classification: dendrite or background. The two 3D CNNs differ by the operations performed by the convolutional blocks (blue arrows in Figure 2a). The convolutional block of 3D U-Net is shown in Figure 2b. The 3D architecture is similar to the 2D U-Net, except that it uses 3D convolutions and poolings rather than their 2D counterparts so that it can exploit the contextual information from each z-slice. A residual layer is appended as suggested by Zhang et al. [31], which is an additional 1 x 1 x 1 convolution in each convolutional block. The resulting feature map is then pointwise summed with the feature map produced by the two 3 x 3 x 3 convolutions in the block. The use of residual layers has been shown by He et al. [32] to promote easier learning, and therefore higher performance.

We created a 3D version of FCDense as part of this study. Similar to how Jegou et al. extended 2D U-Net to 3D [22], we extended FCDense with 3D convolutions, maxpoolings, and transposed convolutions. The convolutional block of 3D FCDense is shown in Figure 2c. The many short connections between feature maps in FCDense result in a more efficient use of training parameters, as well as improved utilization of features across multiple scales.

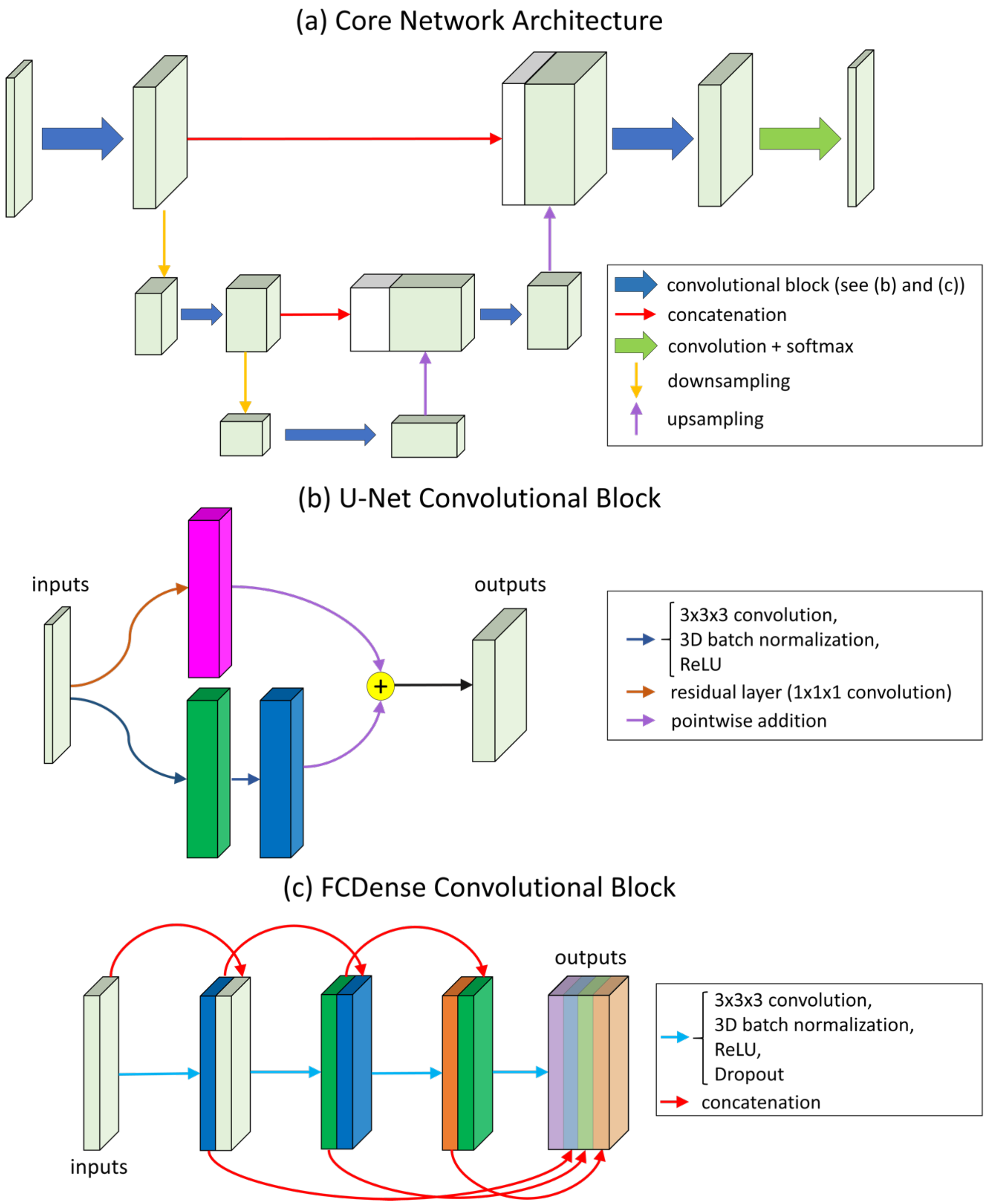

**Figure 2.** (a) Base architecture of 3D U-Net and 3D FCDense. (b) Structure of a convolutional block in the U-Net architecture. (c) Structure of a convolutional block in the FCDense architecture.

We employed image augmentations to create a more varied training dataset, providing a more general training set than would be possible using only the raw initial dataset. We also made various augmentations to the images and corresponding labels before training, including reflections across x-, y-, and z-planes; rotations from +/- 45deg; affine shear transforms from +/- 30deg; and brightness and color contrast adjustments. These are detailed in Table 1.

We define the operation of patching as subdividing reconstructions into smaller volumes. Limiting the image volume is necessary due to memory constraints. We attempted to train 3D CNNs on volumes of different aspect ratios (such as 100 x 100 x 10 or 12 x 12 x 20), however we found that the architectures performed the best with volumes that had 4 z-slices. Thus prior to training, we patched each reconstruction into volumes with a predefined size ranging from (200 x 200 x 4) to (852 x 852 x 4).

To maximize the performance of the 2D and both 3D architectures, we performed two-step hyperparameter optimization (HPO) via the Asynchronous Successive Halving Algorithm (ASHA) [33]. The two-step HPO process is shown schematically in Figure 3. ASHA concurrently trains a model on differing hyperparameters, using a bandit algorithm [34] to select the next set of hyperparameters to test. First, we ran 30 HPO trials of each architecture for initial training on SegNet outputs to select a locally optimal combination of patch width, overlap size, learning rate, augmentations, number of epochs, and batch size. Second, the resulting model for each architecture that performed the best on the validation split of HandSeg was then checkpointed for fine-tuning. Afterwards, without further optimizing patch size and batch size, we ran 30 additional HPO trials for each architecture while training on HandSeg. Our final models were the best performers on the HandSeg validation split. These optimal hyperparameters are referred to in Table 1. Two HPO steps were necessary as the optimal hyperparameters are dependent on the dataset. Thus, the optimal hyperparameters for the initial training may not necessarily be equivalent to those for the fine-tuning step. Especially given the small set of images in the HandSeg dataset, poorly chosen hyperparameters may result in overfitting rather than generalization.

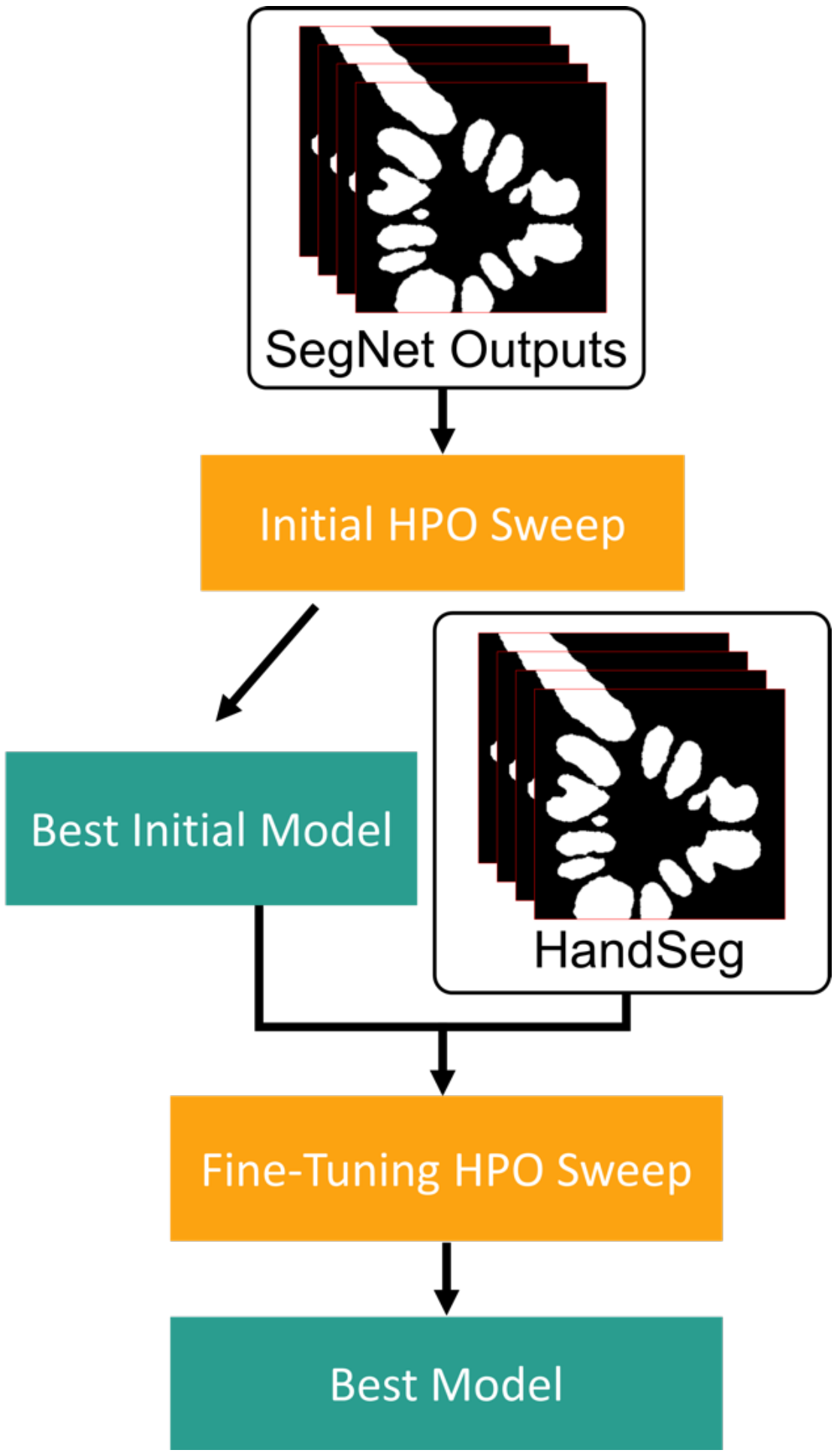

**Figure 3.** Flow diagram of two-step HPO process, starting with HPO training on the SegNet outputs (top) and fine-tuning with HPO training on the hand-segmented data (bottom).

**Table 1.** Hyperparameters selected for each CNN architecture via the two-step HPO process.

| CNN Architecture | 2D U-Net | 3D U-Net | 3D FCDense |
|---|---|---|---|
| Training Patch Volume [voxels] | 776 x 776 x 1 | 456 x 456 x 4 | 560 x 560 x 4 |
| SegNet Outputs Learning Rate | $1.71 \times 10^{-4}$ | $1.07 \times 10^{-4}$ | $6.19 \times 10^{-5}$ |
| SegNet Outputs Epochs | 5 | 4 | 6 |
| HandSeg Learning Rate | $3.80 \times 10^{-4}$ | $3.12 \times 10^{-6}$ | $1.96 \times 10^{-7}$ |
| HandSeg Epochs | 8 | 13 | 11 |
| Batch Size | 16 | 8 | 8 |

Models were trained on the Argonne Leadership Computing Facility (ALCF) Theta GPU cluster, consisting of eight NVIDIA A100 GPUs per node. HPO runs were performed on a single node, with each A100 GPU performing a trial in parallel for each model architecture. The initial HPO step of 3D U-Net took 8 hours and 23 minutes, while training 3D FCDense took 10 hours and 43 minutes. The finetune HPO step took 2 hours and 14 minutes and 2 hours and 56 minutes respectively.

Segmenting an 852 x 852 x 250 voxel 3D reconstruction using a 3D CNN takes ~1 minute using our computational hardware. This translates to a ~5,000x speed improvement compared to manual segmentation, and a ~31x speed improvement compared to the 2D SegNet CNN by Stan et al. [6] which used CPU-based hardware. With respect to data availability, the training images, ground truth volumes, and final 3D segmentations [35] are available for download from the Materials Data Facility [36]. With respect to model availability for reproducibility and testing, the CNNs trained in this study [37,38] can be downloaded and queried via DLHub [39].

### 3.4 Metrics and Evaluation

We used four metrics to evaluate and compare the performance of the different models: accuracy, Intersection over Union (IoU) [40], Boundary F1 Score (BF1) [41], and Boundary Displacement Error (BDE) [42]. Accuracy is the fraction of image pixels that are correctly classified. IoU measures the ratio of the intersection and union of predictions and ground truth, effectively measuring overlap. BF1 and BDE are both boundary-based metrics. BF1 measures the harmonic mean of the proportion of ground truth boundary pixels within four pixels of the predictions, and the proportion of predicted boundary pixels within four pixels of the ground truth, while BDE measures the average distance between predicted and ground truth boundary pixels. These four metrics together give a quantitative measure of both overall segmentation performance and boundary specific accuracy.

## 4. Results

After training and fine-tuning with HPO, we evaluated the models on the portion of the HandSeg dataset that was reserved for testing. We then compared each z-slice in the full segmentation against the ground truth by calculating pixel accuracies, dendrite IoU, BF1, and BDE scores.

Table 2 contains the segmentation results obtained when this testing process was applied to the 2D SegNet from previous literature [7], our three CNNs trained without HPO or fine-tuning, and our three CNNs when optimized via our two-step process. The HPO and fine-tuning procedures clearly further improved the segmentation results. The overall two best CNNs obtained in this study both contain 3D architectures. The best performing CNN according to the metrics is the optimized 3D U-Net which obtained 99.84% accuracy, 95.56% IoU, 98.59% BF1, and BDE of 0.58 pixels. However, the 3D FCDense produced comparable results, but also yielded smoother solid-liquid boundaries which are often desired in solidification characterization experiments (details in later sections).

**Table 2.** Architecture and segmentation metrics (accuracy, intersection over union and boundary F1 score, boundary displacement error) for the CNNs in Section 3.2 and Figure 4. The best scores are in bold.

| CNN Architecture | Acc [%] | IoU [%] | BF1 [%] | BDE [pixels] |
|---|---|---|---|---|
| 2D SegNet (previous literature) | 99.67 | 91.14 | 97.14 | 1.13 |
| 2D U-Net (no HPO or fine-tuning) | 99.54 | 88.30 | 97.36 | 1.42 |
| 3D U-Net (no HPO or fine-tuning) | 99.56 | 88.72 | 97.45 | 1.37 |
| 3D FCDense (no HPO or fine-tuning) | 99.57 | 88.84 | 97.05 | 1.55 |
| 2D U-Net with HPO and fine-tuning | 99.82 | 94.88 | **98.68** | 0.70 |
| 3D U-Net with HPO and fine-tuning | **99.84** | **95.56** | 98.59 | **0.58** |
| 3D FCDense with HPO and fine-tuning | 99.80 | 94.37 | 98.66 | 0.68 |

While the quantitative metrics provide an important and simple way to assess CNN performance, the conclusions that can be drawn from them are limited. Pixelwise accuracy already starts with a high benchmark of 99.67% due to the large number of background pixels in the images, and is only increased to 99.84% by our best model. Furthermore, pixelwise accuracy does not reveal information about the location of improperly segmented regions. While the boundary-based metrics provide important insights about performance at dendrite body boundaries, the BF1 score alone does not reveal much about boundary jaggedness or incorrectly segmented interfaces, due to its four-pixel tolerance. Additionally, these metrics act only on z-slices, rather than on entire volumes. Figure 4 shows a HandSeg z-slice reserved for testing, as well as overlays generated by using the original 2D SegNet [7] and our three best-performing CNNs with HPO and fine-tuning. The 2D SegNet overlay contains many green pixels around the dendrite bodies (indicating over-segmentation), and connects two dendritic sections that should have been separated (red arrow). Our three optimized CNNs all segment the reconstructed image with higher boundary accuracy, as seen in Figure 4 and are consistent with the quantitative measurements in Table 2. Furthermore, none of the three optimized CNNs incorrectly connects the dendritic bodies.

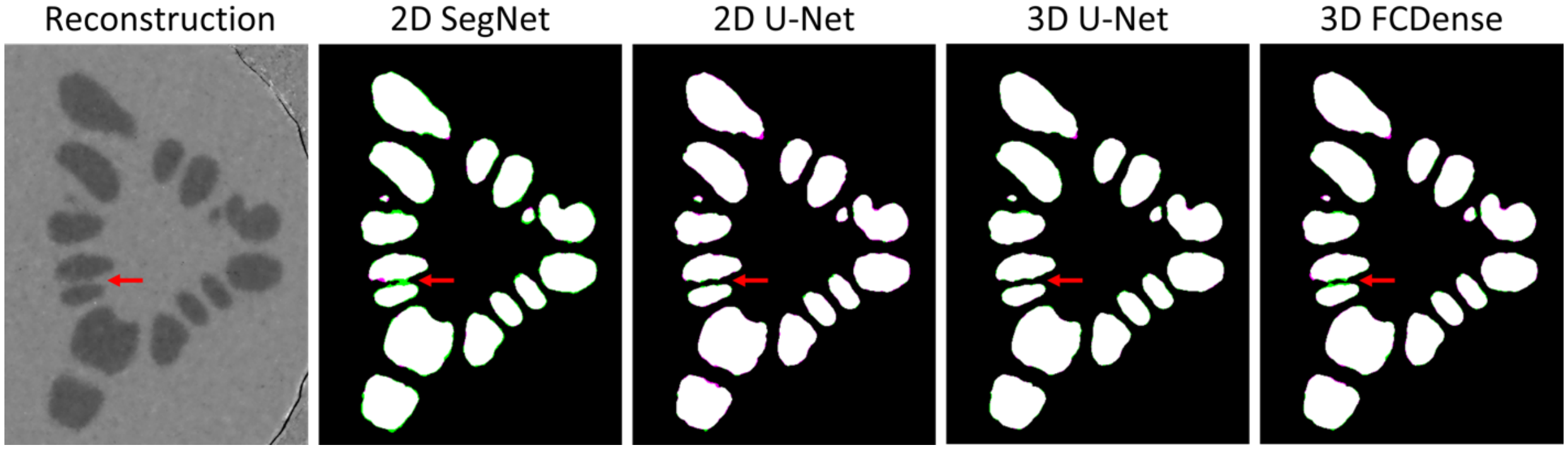

**Figure 4.** (Left to Right): Raw image z-slice of dendrite body. 2D SegNet segmented result, overlaid with ground truth results. 2D U-Net segmented result, overlaid with ground truth results. 3D U-Net segmented result, overlaid with ground truth results. 3D FCDense segmented result, overlaid with ground truth results.

We also examined the outputs of each network qualitatively via x-y plane overlays and x-z plane overlays (both perpendicular to the z-slices). This allowed us to identify networks that inaccurately connect interfaces between dendrite bodies or create jagged boundaries.

Figures 5 and 6 demonstrate the success of the networks in creating smooth segmentations across z-slices as well. In Figure 5, the segmentation created by the 2D SegNet shows an over-abundance of false-positive voxels, while the 2D U-Net, 3D U-Net, and 3D FCDense match the ground truth more closely. In the region denoted by a red arrow in Figure 5, the segmentations generated by the optimized networks are more consistent and do not create a jagged, over-segmented edge where there is ambiguity in the original volume.

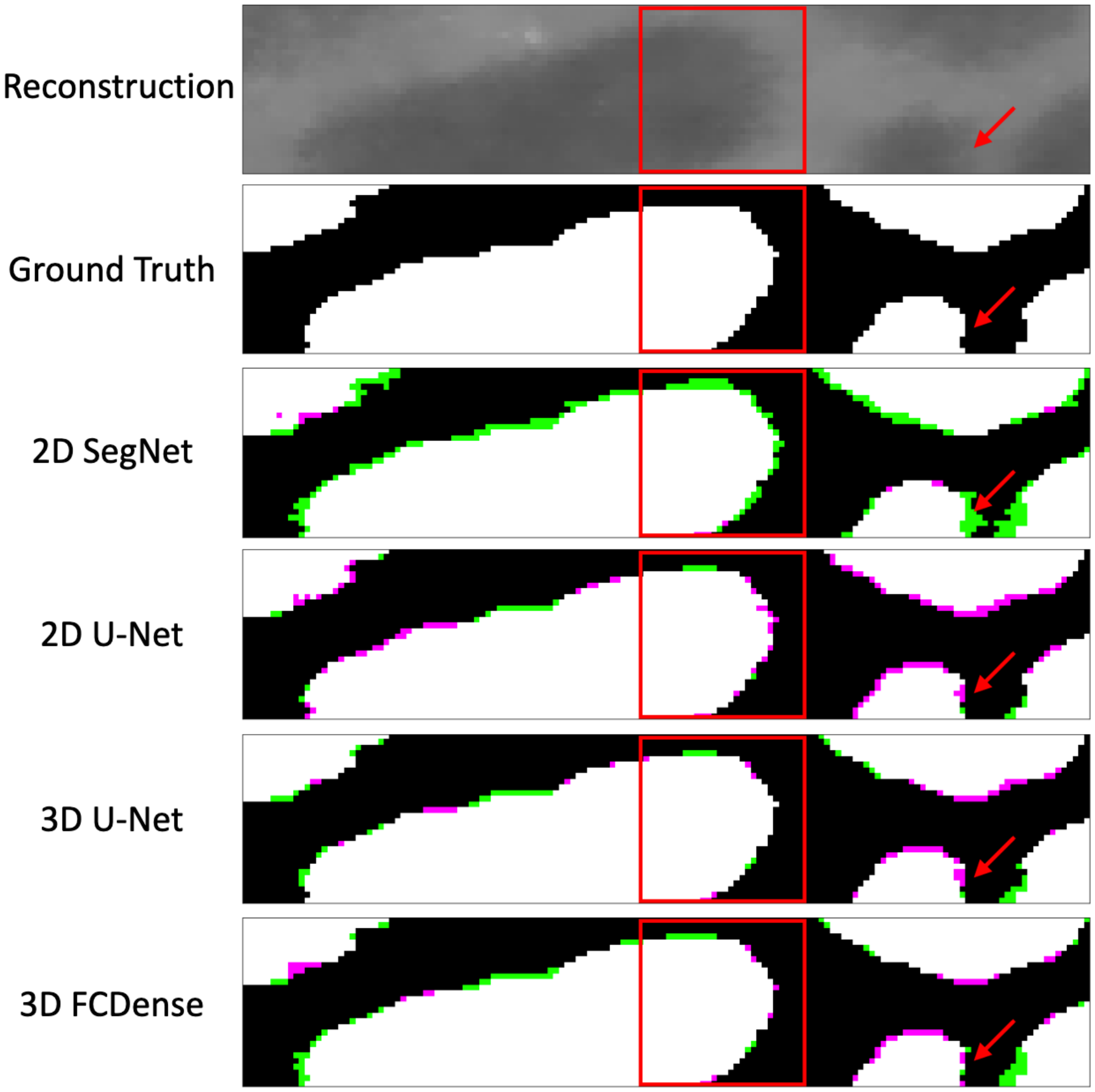

**Figure 5.** (Top to Bottom): Raw cross-section image across slices in the z-axis of dendrite body. Original 2D SegNet cross-section result, overlaid with manual ground truth. 2D U-Net cross-section result, overlaid with manual ground truth. 3D U-Net cross-section result, overlaid with manual ground truth. 3D FCDense cross-section result, overlaid with manual ground truth.

The region marked with a red box in Figure 5 is reproduced at higher magnification in Figure 6. These results highlight the ability of the 3D neural networks to generate smoother dendrite-background boundaries by using contextual information from adjacent layers. While the 2D SegNet and the optimized 2D U-Net both generated jagged edges in their segmentation, the 3D U-Net and 3D FCDense created a smoother edge with higher accuracy compared to the ground truth and better reflecting the properties of the dendrite. Although 3D U-Net contained the best overall segmentation metrics (Table 2), the smoother boundary produced by 3D FCDense is often desired when analyzing in-situ experimental datasets of dendritic growth.

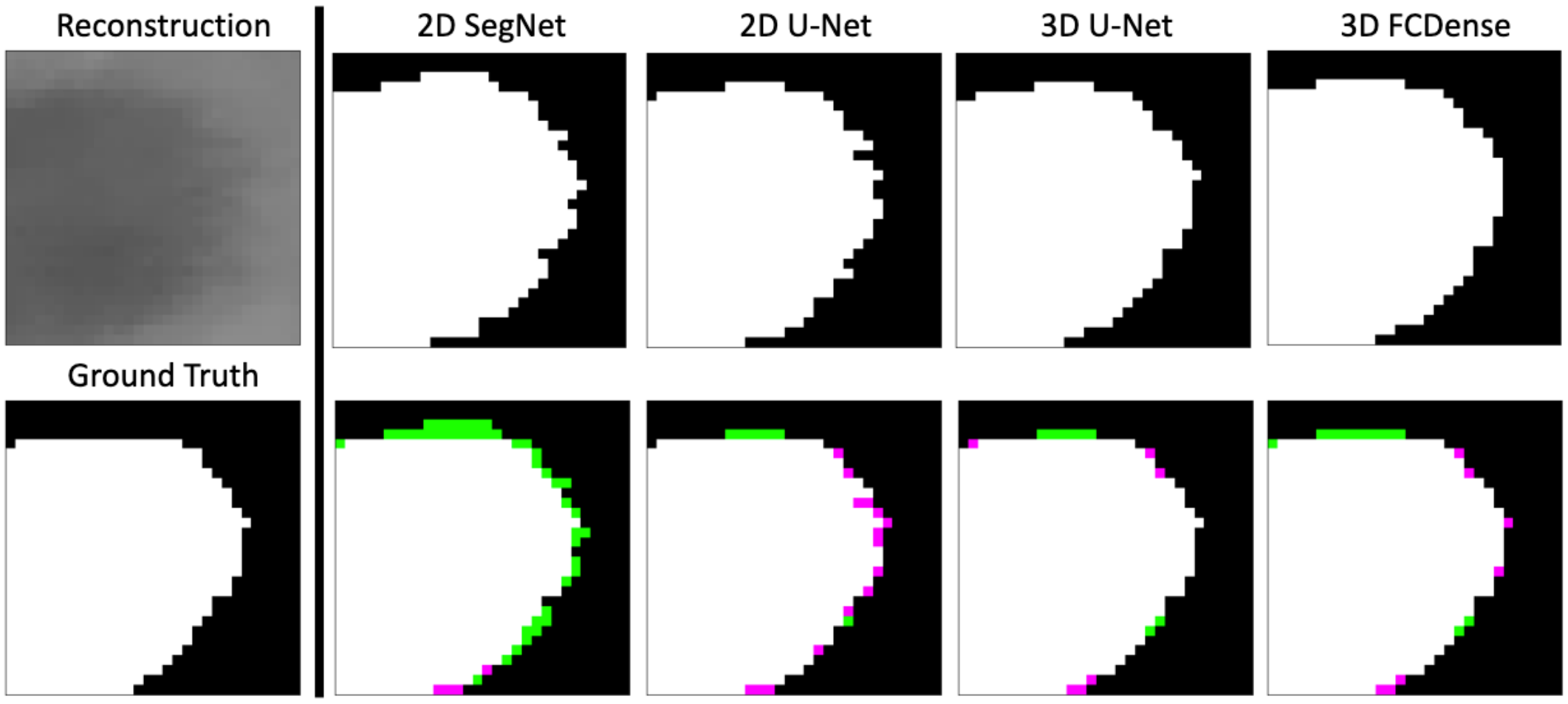

**Figure 6.** (Left to Right): Original and ground truth slices across x- and z- axes. 2D SegNet segmentations (original and overlaid with ground truth). 2D U-Net segmentations (original and overlaid with ground truth). 3D U-Net segmentations (original and overlaid with ground truth). 3D FCDense segmentations (original and overlaid with ground truth).

## 5. Discussion

The optimized 2D U-Net, 3D U-Net, and 3D FCDense architectures greatly outperformed the classical computer vision benchmarks (Figure 1) as well as the previous state-of-the-art 2D SegNet across all qualitative and qualitative measures, thereby demonstrating their superior performance in segmentation continuity and boundary accuracy. These advantages are showcased in the z-slice overlays in Figure 4, where all three optimized networks correctly separated all dendrite bodies, while the 2D SegNet incorrectly connected two dendrites, illustrating the importance of 3D segmentation.

Both 3D networks and the optimized 2D U-Net have comparably high pixel accuracies and BF1 scores. Similarly, 3D FCDense and 2D U-Net have nearly identical dendrite IoUs, indicating similar segmentation continuity, with 3D U-Net performing approximately 1% better than the other two models. 3D U-Net has much lower BDE than 2D U-Net, implying less variation across slices and therefore smoother boundaries. The superior performance of the 3D architectures is further supported by the overlays shown in Figures 5 and 6. Both the 2D SegNet and the optimized 2D U-Net produced jagged protrusions and cuts that are not present on the 3D model outputs.

The ability of 3D CNNs to output smoother boundaries in the z-direction than 2D CNNs is largely due to the inherent CNN architecture, and the training data type. A 2D CNN uses a series of 3x3x1 convolutional layers that probe the training data one image at a time; thus, it has no information regarding the data in the z-slices above and below the current slice. By contrast, the 3D CNN architectures contain 3x3x3 convolutional layers that probe three times more data than the 2D architecture. Furthermore, the 3D CNNs are trained on four z-slices at a time, and so are able to learn more information about the 3D nature of the XCT experimental data. The 3D U-Net performed the best according to the segmentation metrics in Table 2, but 3D FCDense had smoother boundaries and performed the best according to visual inspection (such as the x-slice in Figure 6). The objective versus subjective discrepancy between these networks is viewed as an advantage since networks may be tuned depending on the goals of the scientific research. In the case where an input dataset (such as the 3D grayscale reconstructions used in this study) and the associated ground truth training dataset are both highly accurate, then using a 3D U-Net would be ideal because it most accurately reproduces the high-quality ground truth segmentations. However, if the input dataset has poor quality and the ground truth dataset has errors, then a CNN that reproduces the poor segmentations may not be desired. In the latter case, it is an advantage that the 3D FCDense architecture did not reproduce the ground truth perfectly but instead output segmentations with smoother boundaries (which are highly desired in XCT studies of solidification). Thus, both 3D U-Net and 3D FCDense can be viewed as successful, depending on the application.

The results furthermore indicate the need for both fine-tuning and HPO to create a model that supersedes the previous state of the art. Without fine-tuning the models could not produce accurate outputs, and models without HPO would lead to sub-optimal local minima, creating less accurate outputs. These conclusions are validated quantitatively in Table 2. With both HPO and fine-tuning, all three networks can generate a segmentation that outperforms the previous state of the art.

We observed that some hyperparameters had a greater impact on model performance than others. Specifically, the learning rate and number of slices each have a large impact on model performance. Interestingly, we determined that a depth of four slices is optimal in the z-direction. We hypothesize that a larger number of slices creates too large a context for the network to perform well in all slices. We note that while adjusting the number of z-slices resulted in large changes in model performance, adjusting the patch size (that is, context in the x- and y-axes) did not yield significant changes in model performance across volumes.

We applied the trained 3D U-Net architecture to one of the reconstructions from the XCT experiment. We show the resulting mesh in Figure 7. The mesh was left unsmoothed to highlight the 3D CNN performance. The microstructure contains ten primary arms (some with secondary arms) which protrude from a common dendritic root. A more detailed analysis of this dataset will be presented in a future publication.

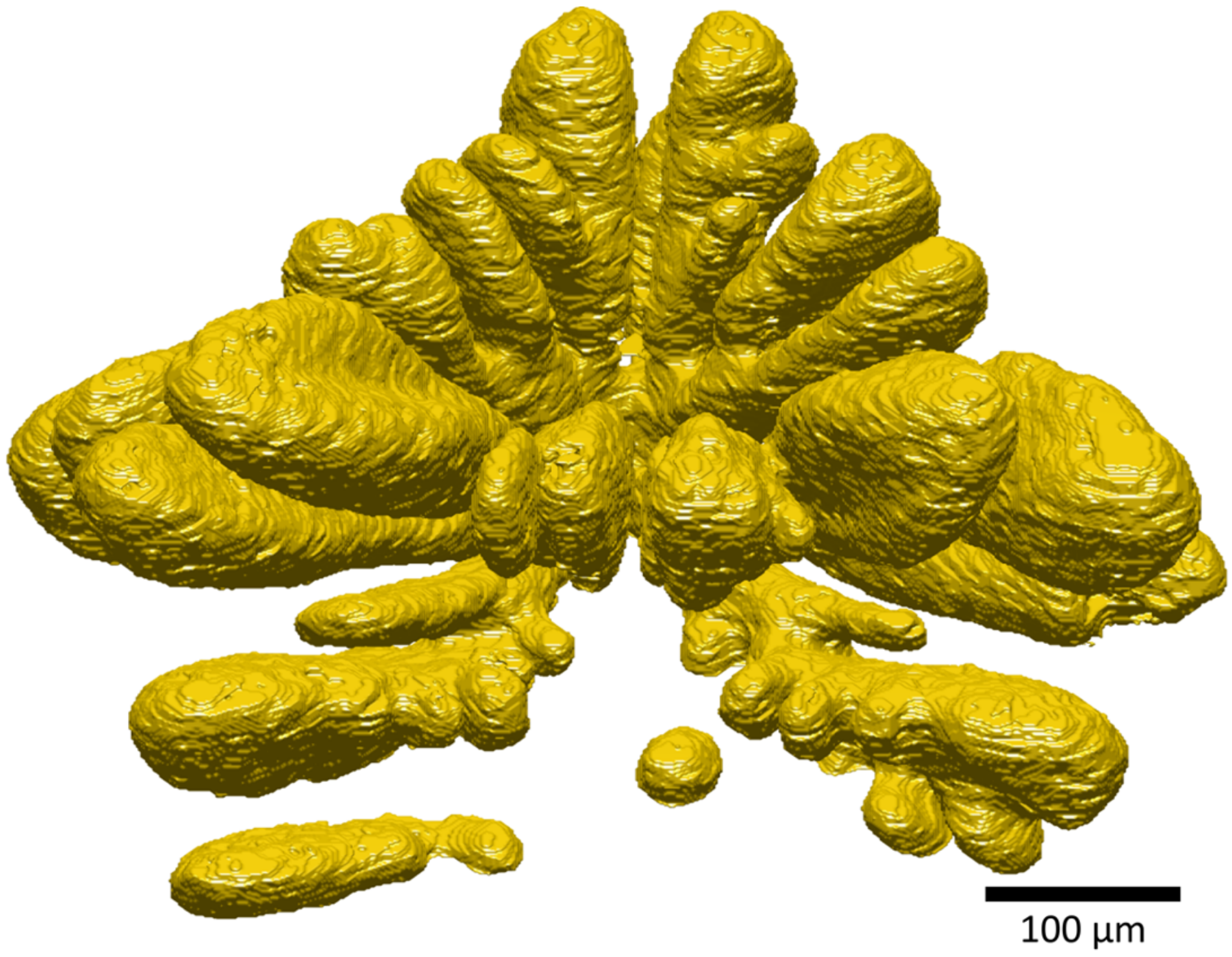

**Figure 7.** An unsmoothed 3D rendering of the entire dendritic morphology as segmented by 3D U-Net.

Future work includes using different loss functions within existing CNN architectures to improve the efficiency of our models and the accuracy of the generated segmentations. Since the most inaccurate areas in these models are located where dendrite boundaries are close together, a loss function focused on boundary performance, such as the already-existing Active Boundary Loss [43] or a novel loss function, may enable greater performance. Additionally, to reduce the time needed for manual segmentation even further, we intend to use 3D phase field simulations [44] to generate both training data and ground truths computationally, without the need for human intervention. A 2D version of this approach has been shown to be successful by Yeom et al. [45]. We are hopeful that a 3D version can be developed for segmentation of large XCT and serial sectioning datasets.

## 6. Conclusions

This study highlights the superior ability of 3D CNNs to segment 3D datasets. A 3D XCT dataset of dendritic solidification in an Al-Zn alloy was used to train three CNNs (2D U-Net, 3D U-Net, and 3D FCDense). To the best of our knowledge, this is the first time that a 2D FCDense architecture has been extended to 3D. All three architectures achieved improved segmentation performance after fine-tuning and HPO. The optimized 2D U-Net performed better than past 2D benchmarks, while the 3D U-Net and 3D FCDense networks achieved the best segmentations. Both 3D CNNs achieved smoother boundaries than any of the 2D CNNs. The 3D U-Net produced the best segmentations according to quantitative metrics, with an accuracy of 99.84%, an IoU of 95.56%, a BF1 score of 98.59%, and a BDE score of 0.58 pixels, while 3D FCDense produced the smoothest boundaries and best segmentations according to visual inspection. By showing how 3D CNNs can be used to generate highly accurate segmentations of large 3D image data in only minutes, this work demonstrates how deep learning can be leveraged to enable deeper understanding of phase transformations.

## 7. Acknowledgements

The authors thank Bo Lei (Carnegie Mellon University), Elizabeth Holm (Carnegie Mellon University), Joshua Pritz (Northwestern University), Marta Garcia Martinez (Argonne National Laboratory), and Aniket Tekawade (Argonne National Laboratory) for guidance and fruitful discussions. This work was supported in part by the U.S. Department of Commerce, National Institute of Standards and Technology as part of the Center for Hierarchical Materials Design (CHiMaD) under award number 70NANB19H005, the KAIST-funded Global Singularity Research Program for 2021 and 2022 under award number 1711100689, and the U.S. Department of Energy, Office of Science, under contract DE-AC02-06CH11357. This research used resources of the Argonne Leadership Computing Facility, which is a DOE Office of Science User Facility supported under Contract DE-AC02-06CH11357. The authors also acknowledge the University of Chicago Research Computing Center for support of this work.